\pdfoutput=1
\documentclass[11pt,a4paper]{article}
\usepackage{acl}

\usepackage[most]{tcolorbox}

\usepackage{times}
\usepackage{latexsym}
\usepackage{graphicx}
\usepackage{booktabs}
\usepackage[T1]{fontenc}
\usepackage[utf8]{inputenc}

\usepackage{microtype}

\usepackage{inconsolata}

\usepackage{graphicx}

\usepackage{dblfloatfix} % bertie
\usepackage{pifont} % bertie
\usepackage{array} % bertie
\usepackage{makecell} % bertie
\usepackage{ragged2e} % bertie - for \RaggedRight
\usepackage[table]{xcolor} % bertie
\usepackage{rotating}
\usepackage{tabularx}
\usepackage{colortbl}
\usepackage{multirow}
\usepackage{enumerate}
\usepackage{longtable}
\usepackage{booktabs}       % professional-quality tables
\usepackage{hyperref}       % hyperlinks
\usepackage{url}            % simple URL typesetting
\usepackage{pdflscape}      % landscape for wide tables
\usepackage{float}          % Better float control with H option
\usepackage{tcolorbox}

\usepackage{todonotes}

\newcolumntype{P}[1]{>{\centering\arraybackslash}p{#1}}
\newcolumntype{M}[1]{>{\centering\arraybackslash}m{#1}}
\newcolumntype{L}[1]{>{\raggedright\arraybackslash}m{#1}}
\newcolumntype{C}[1]{>{\centering\arraybackslash}m{#1}}
\definecolor{DarkGreen}{RGB}{0,100,0}   % a standard dark green
\definecolor{ForestGreen}{RGB}{34,180,34}
\definecolor{PeachPuff}{RGB}{255,218,185}
\definecolor{Maroon}{RGB}{128,0,0}
\definecolor{IndianRed}{RGB}{205,92,92}
\definecolor{LightPink}{RGB}{255,182,193}

\newcommand{\vcolorbox}[3][1.4em]{%
  \colorbox{#2}{\makebox[#1][c]{#3}}%
}

\title{The AI Consumer Index (ACE)}

\author{
    \textbf{Julien Benchek}$^{1, \wedge}$ \quad
    \textbf{Rohit Shetty}$^{1, \wedge}$ \quad
    \textbf{Benjamin Hunsberger}$^{1}$ \quad
    \textbf{Ajay Arun}$^{1}$ \quad \\
    \textbf{Zach Richards}$^{1}$ \quad
    \textbf{Brendan Foody}$^{1}$ \quad
    \textbf{Osvald Nitski}$^{1}$ \quad
    \textbf{Bertie Vidgen}$^{1}$\thanks{Email: apex@mercor.com} \\
    $^1$Mercor \quad $^\wedge$Joint first authors
}

\begin{document}
\pagestyle{plain}
\thispagestyle{plain}

\maketitle
\begin{abstract}
We introduce the first version of the \textbf{AI Consumer Index} (ACE), a benchmark for assessing whether frontier AI models can perform everyday consumer tasks. ACE contains a hidden heldout set of $400$ test cases, split across four consumer activities: shopping, food, gaming, and DIY. We are also open sourcing $80$ cases as a devset with a CC-BY license. For the ACE leaderboard we evaluated 10 frontier models (with web search turned on) using a novel grading methodology that dynamically checks whether relevant parts of the response are grounded in the retrieved web sources. GPT 5 (Thinking = High) is the top-performing model, scoring $56.1\%$, followed by o3 Pro (Thinking = On) at $55.2\%$ and GPT 5.1 (Thinking = High) at $55.1\%$. Model scores differ across domains, and in Shopping the top model scores under $50\%$. We find that models are prone to hallucinating key information, such as prices. ACE shows a substantial gap between the performance of even the best models and consumers' AI needs.
\end{abstract}

\setlength{\parindent}{0pt}
\section{Introduction}

\begin{figure}[t]
\centering
\includegraphics[width=1.05\linewidth]{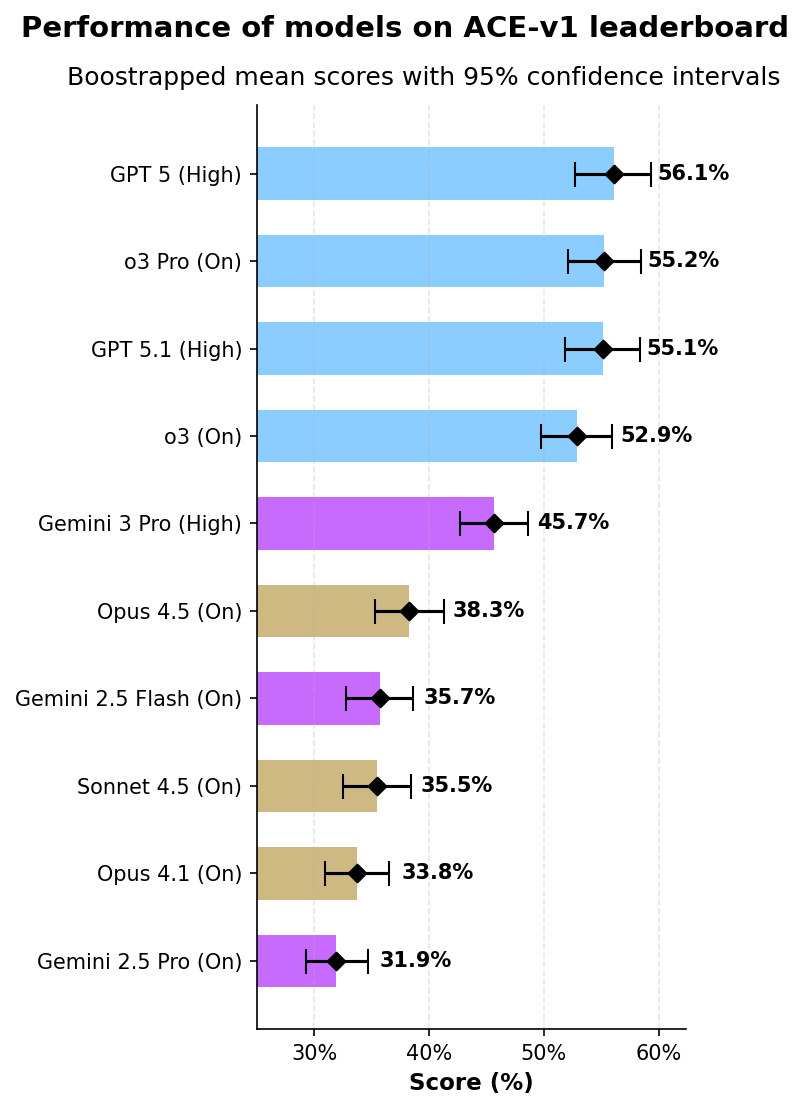}
\caption{The ACE leaderboard (\textbf{ACE-v1-heldout}).}
\label{fig:model-overall-mean-score}
\end{figure}

Consumer use of AI is widespread and accelerating. A report in June 2025 from Menlo Ventures found that $61\%$ of American adults had used AI in the previous six months and $19\%$ used it every day \citep{Benchek2025State}. In September 2025, OpenAI reported that daily ChatGPT messages had increased from 451 million in June 2024 to 2,627 million in June 2025 \citep{Chatterji2025How}. Consumer use drove much of this growth, with the proportion of non-work-related messages increasing from $53\%$ to $73\%$. Bain \& Co estimate that consumer spending on AI is $\$12$ billion per year, with substantial growth potential as $97\%$ of consumers use only free versions of AI products \citep{SommerfeldGriffin2025FiveTypesAIConsumers}. At the same time, numerous studies show that the public is concerned about the accuracy and trustworthiness of AI models and products \citep{McClain2025USPublicAIExperts, ConsumerReports2024AIAlgorithms, Benchek2025State, WEF2025ConsumersAI}. Existing benchmarks have not paid enough attention to consumer applications of AI, instead focusing on abstract reasoning capabilities or, to a lesser extent, professional work \citep{vidgen2025aiproductivityindexapex} and coding \citep{jimenez2024swebenchlanguagemodelsresolve, aleithan2024swebenchenhancedcodingbenchmark, ma2025swefficiencylanguagemodelsoptimize}. To tackle this problem, we are releasing the AI Consumer Index (ACE), a benchmark that assesses whether AI models can meet the everyday needs of consumers.\footnote{\href{ehttps://www.mercor.com/apex/ace-leaderboard/}{mercor.com/ace-leaderboard}}
\vspace{1em}

%ACE evaluates one of the most important applications of frontier AI systems -- helping people with everyday activities 
%Existing benchmarks have largely ignored these commercially important but more nuanced and subjective uses of AI.
% The Benchmark was constructed in three steps.
% First, we sourced experts with relevant experience in each domain.
% Second, experts created prompts that match real, sophisticated requests.
% Third, experts created quality rubrics for evaluating model responses.

\begin{figure*}[!h]
\centering
\includegraphics[width=0.95\linewidth]{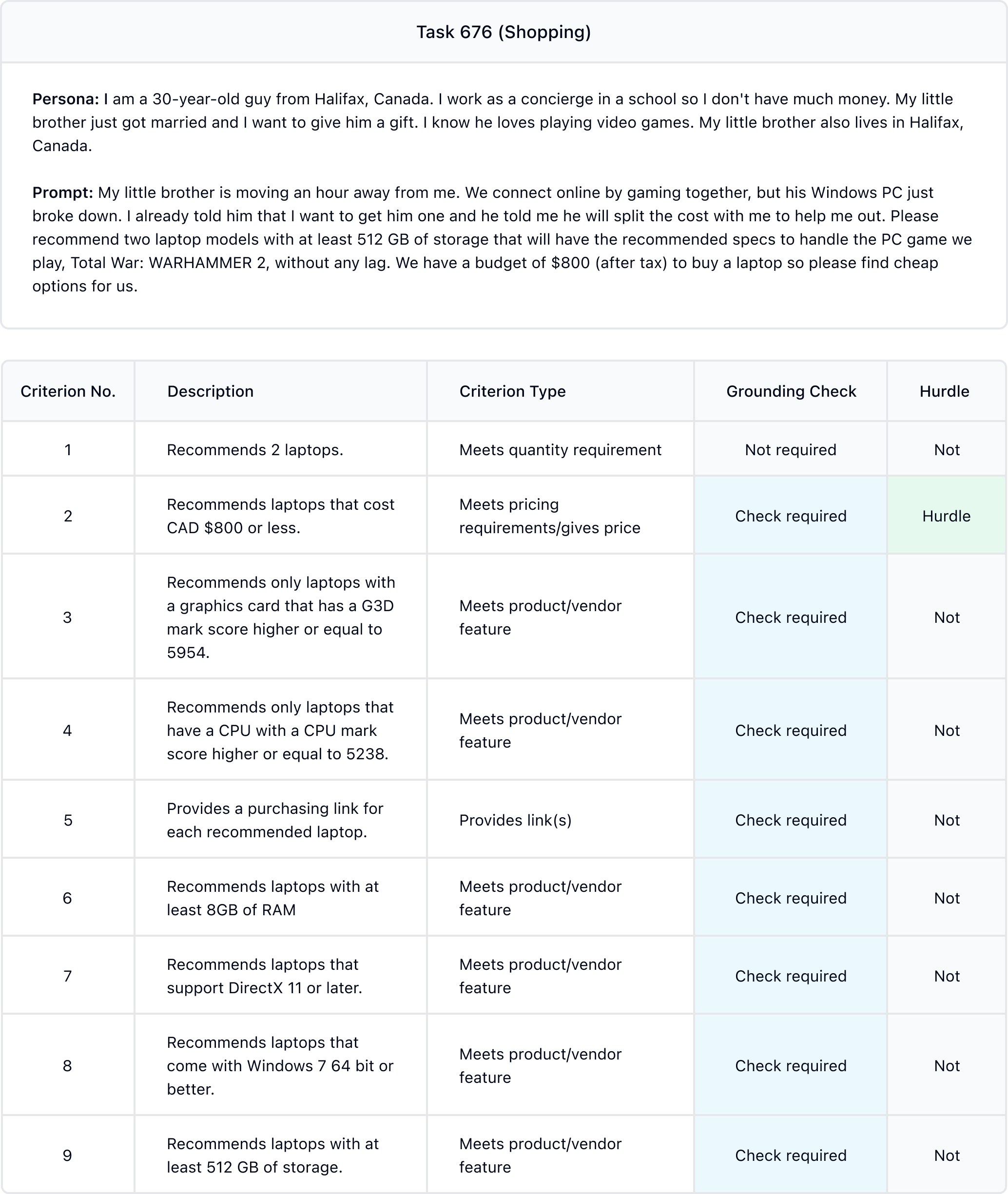}
\caption{Example rubric for \textbf{Shopping (ID 676)} with 9 criteria. This case is from \textbf{ACE-v1-dev} and is not used in the ACE leaderboard.} \label{fig:example-rubric}
\end{figure*}

\begin{figure*}[htbp]
\centering
\includegraphics[width=0.9\linewidth]{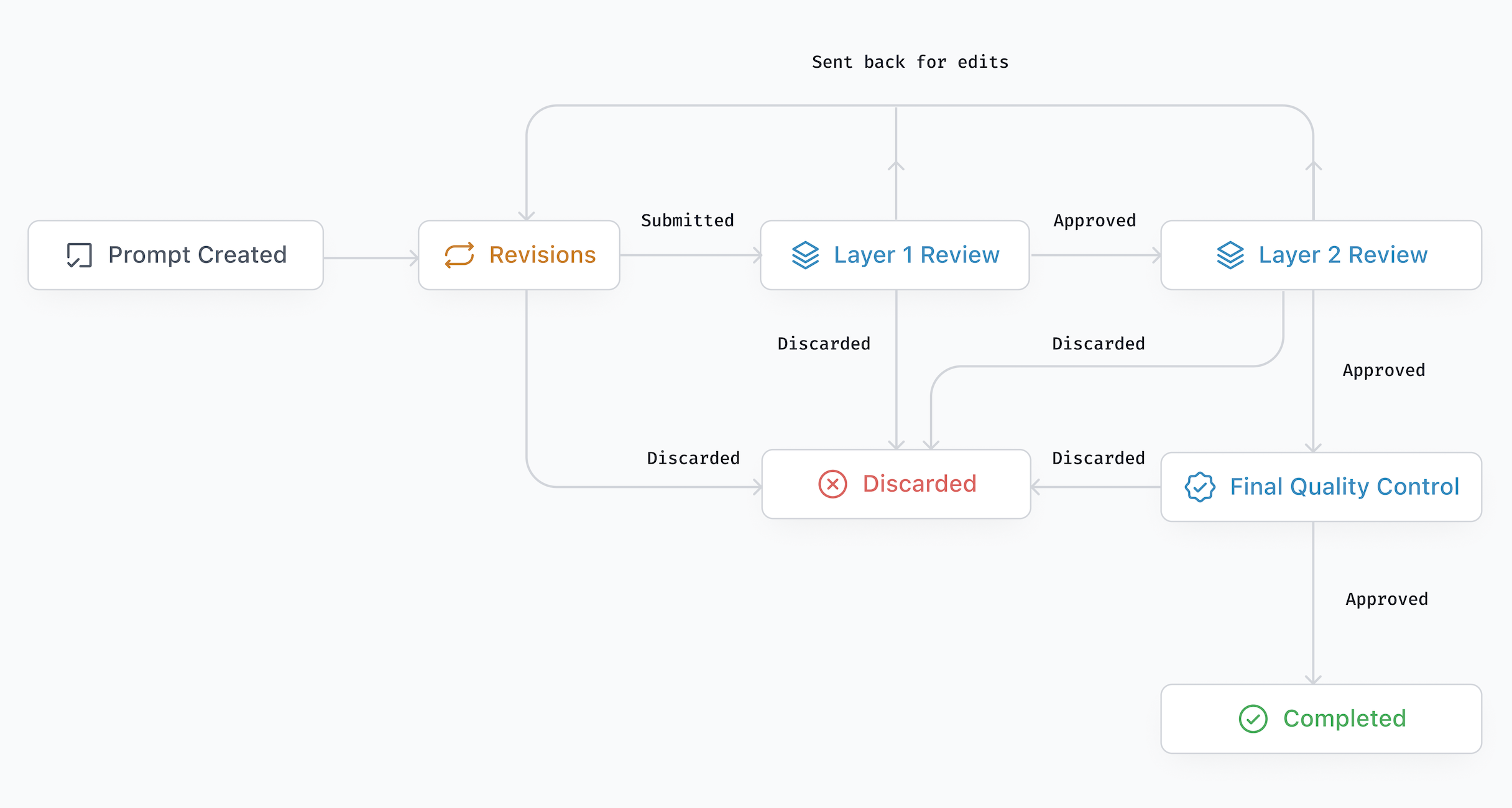}
\caption{Overview of the production process for creating cases in the AI Consumer Index. Quality control is applied at every step.}
\label{fig:model-splash}
\end{figure*}

ACE contains a heldout set of $400$ tasks, which we call \textbf{ACE-v1-heldout}. It is hidden to minimize the risk of contamination and overfitting. The tasks are evenly divided across four domains of consumer activity: (1) Shopping, (2) DIY, (3) Gaming, and (4) Food, as described in Table~\ref{tab:dataset_overview}. To advance open research, we are open sourcing 20 cases from each domain (with prompts, metadata and grading rubrics), comprising 80 cases total.\footnote{\href{https://huggingface.co/datasets/mercor/ACE}{huggingface.co/datasets/mercor/ace}} We call this \textbf{ACE-v1-dev}. An example prompt and rubric is given in Figure~\ref{fig:example-rubric}.
We are also making our eval harness open source for full reproducibility.\footnote{\href{https://github.com/Mercor-Intelligence/apex-evals}{github.com/Mercor-Intelligence/apex-evals}}
The public-facing leaderboard for ACE is initially released with results for 10 models. %To minimize noise, we collect model responses eight times for each prompt and use the mean score. All scores are independently graded by a judge LM.
\vspace{1em}

% The rubrics in ACE-v1-heldout have finegrained labels for each criterion that enable high-fidelity loss analysis: (1) whether the criterion is grounded or not (i.e., requires the model to make a claim based off information in a retrieved web source) and (2) using a newly developed taxonomy of criteria, the criterion type (i.e., meeting a requested quantity, meeting a product feature, or returning a link). 
% Using these labels, we show that many models are worse at grounding their responses than meeting the requirements of the prompt -- they are prone to making up information or providing dead links in order to satisfy the request. We also show that models perform worse at meeting nuanced requests, such as compatibility requirements in Gaming (most models score under 40\%) or providing suitable safety warnings in DIY (most models score under 50\%). 
% \vspace{1em}

% We have designed ACE-v1.0 so the evaluation is ``timeless'' -- even though it focuses on web search in consumer applications, which by its nature is changing constantly. 
% We use the rubrics to autograde responses using a panel of judge LMs (see Appendix A).

\begin{table}
\centering
\small
\caption{Overview of the \textbf{ACE-v1-heldout} and \textbf{ACE-v1-dev} datasets, showing the number of tasks, the average number of criteria per domain, the average number of hurdles, and the percentage of criteria that are grounded.}
\label{tab:dataset_overview}
\begin{tabular}{L{2.15cm}|C{0.6cm} C{0.8cm} C{0.7cm} C{0.7cm}}

\toprule
\textbf{Domain} & \textbf{Tasks} & \makecell{\textbf{Avg}\\\textbf{criteria}} & \makecell{\textbf{Avg}\\\textbf{hurdles}}  & \makecell{\textbf{Grnd}\\\textbf{crit}} \\
\midrule
\multicolumn{5}{l}{ACE-v1-heldout} \\  
\midrule
DIY & 100 &  10.71 & 1.01 & 0.00 \\
Food & 100 & 7.65 & 1.67 & 0.00 \\
Gaming & 100 & 5.41 & 1.35 & 42\% \\
Shopping & 100 & 5.21 & 1.25 & 74\% \\
\textbf{Unweighted avg} & \textbf{100} & \textbf{7.25} & \textbf{1.32} & \textbf{29\%} \\

\midrule
\multicolumn{5}{l}{ACE-v1-dev} \\  
\midrule
DIY & 20 &  10.50 & 1.05 & 0.00 \\
Food & 20 & 7.50 & 1.70 & 0.00 \\
Gaming & 20 & 5.35 & 1.15 & 26\% \\
Shopping & 20 & 6.25 & 1.35 & 78\% \\
\textbf{Unweighted avg} & \textbf{20} & \textbf{7.40} & \textbf{1.31} & \textbf{26\%} \\
\bottomrule
\end{tabular}
\end{table}

\begin{figure*}[t]
\centering
\includegraphics[width=0.9\linewidth]{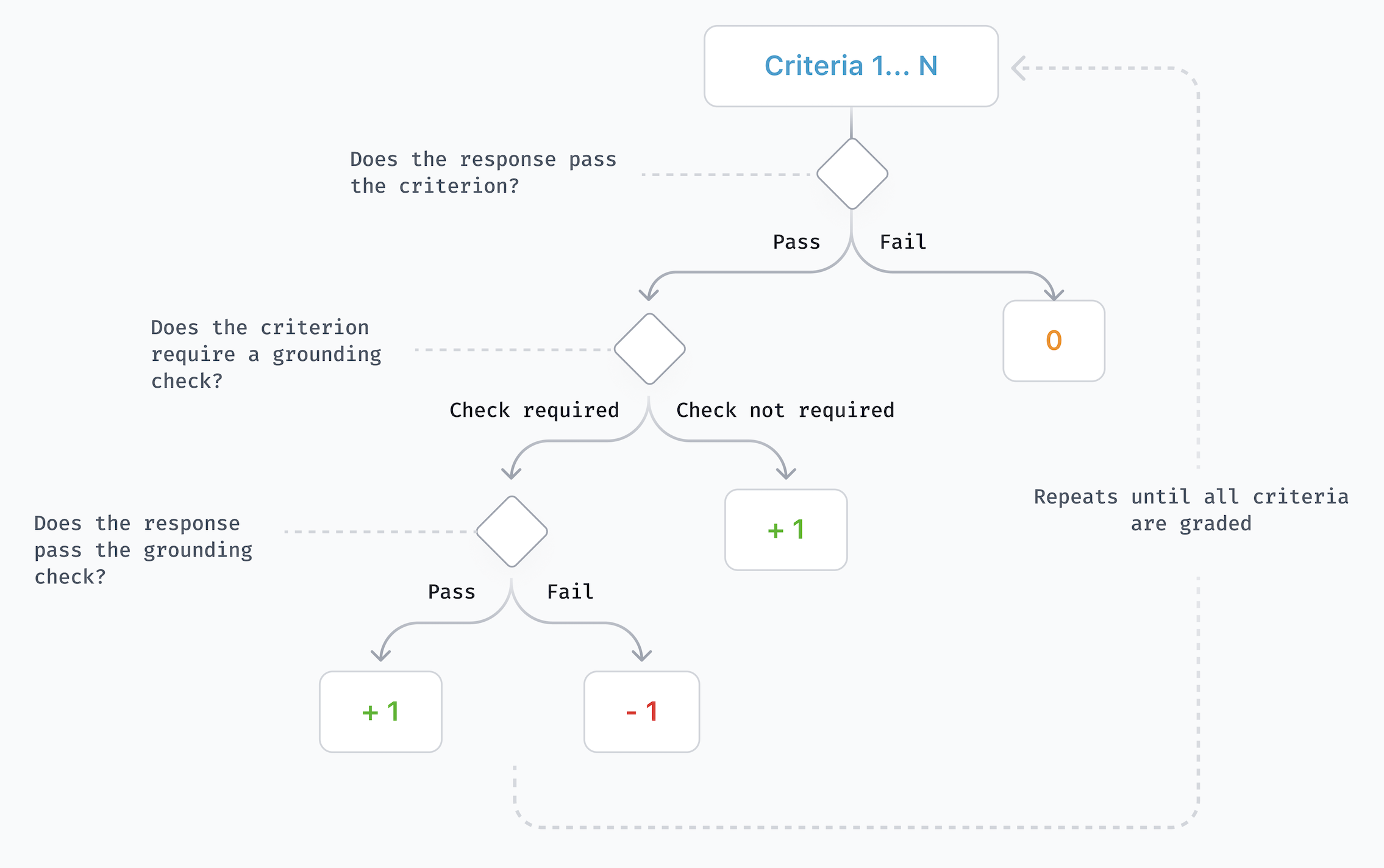}
\caption{Hierarchical process for grading criteria in \textbf{ACE-v1}.}
\label{fig:criteria_grading}
\end{figure*}

\section{Dataset overview}
%ACE-v1-heldout comprises $n = 400$ cases, split evenly across the four consumer domains, as shown in Table~\ref{tab:dataset_overview}. 

\subsection{Experts and quality control}
Each case was created by subject matter experts and reviewed multiple times, as shown in Figure~\ref{fig:model-splash}.
Experts were sourced through the Mercor Platform with appropriate experience for each consumer activity domain, such as personal shoppers, stylists, and shopping magazine editors for Shopping; game developers and professional gamers in Gaming; chefs, food magazine editors, and nutritionists for Food; and tradespeople, construction workers, and mechanical engineers for DIY. 
Throughout the project we continually gave feedback to the experts as we iterated on scope and design. In total, 47 experts contributed at least one case to ACE-v1 (including both ACE-v1-dev and ACE-v1-heldout).

\subsection{Taxonomy of workflows}
For each domain in ACE we developed a taxonomy of workflows to ensure dataset diversity and to better understand common AI consumer use cases. We interviewed experts working on ACE and manually reviewed several rounds of data. There are 5 workflows in Shopping, 2 workflows in DIY, 4 workflows in Gaming, and 3 workflows in Food, as shown in Appendix~\ref{sec:appendix_workflows}, with the number of cases that each workflow accounts for in ACE-v1-heldout.

\subsection{Prompts}
Each prompt contains a \textit{request}, stating a clear objective. Most prompts are also provided with a \textit{persona}, describing the background and primary objective of the user. For the ``Gifting'' workflow in Shopping we provide both a Giver and Recipient persona. Where personas exist, the task can only be successfully executed by taking into account the persona and request together. An example is given in Figure~\ref{fig:example-rubric}.
\vspace{1em}

We experimented with several versions of the prompt phrasing to ensure that users' expectations of a high-quality model output (as codified in the rubric) are fairly communicated to the model. Initially, experts created very simple prompts that did not specify exactly what they wanted. Models did not discern their expectations and so would routinely not return key information they expected, such as a link to purchase an item or the price of the item. To make the ACE leaderboard grading process fair, we append a short piece of text to the end of each prompt that makes the expectations of users explicit. This text is customized to the workflow in each domain, and is given in Appendix~\ref{sec:appendix_prompt_specification}.

\subsection{Criteria}
For each prompt, experts create a rubric of criteria to evaluate the quality of responses.
Each criterion is an objective, specific, and self-contained statement about the response, phrased as a descriptive claim. Each criterion can be assessed as Pass or Fail by a human or LM judge \citep{saadfalcon2024lmunitfinegrainedevaluationnatural, arora2025healthbenchevaluatinglargelanguage, starace2025paperbenchevaluatingaisability}. 
The mean number of criteria for Shopping tasks is $5.21$, Gaming is $5.41$, DIY is $10.71$, and Food is $7.65$. 
Each criterion has two metadata tags that are used in the grading methodology: (1) whether it assesses an aspect of the response that requires grounding or not (see below) and (2) whether it is a ``hurdle'' (see below). We also provide a label for the criteria type, as shown in Table~\ref{tab:criteria_scores}. There are 7 criteria types in DIY, 10 in Food, 8 in Gaming, and 6 in Shopping. A small number of criteria types appear in multiple domains (e.g., ``Provides link(s)'' and ``Other''). 
%We created this taxonomy iteratively by working with experts and reviewing prompts, criteria, and model responses. We identified reoccurring criteria and found that categorizing them helped minimize the risk of unprincipled gaps in the rubric. The fine-grained labels can also be used for loss analysis to understand the strengths and weaknesses of models (see below). 

\section{Experimental setup}
We tested 10 frontier models from Anthropic, Google Deepmind, and OpenAI against ACE-v1-heldout. Responses were collected from the models' respective APIs at the end of November 2025. Thinking is turned on for all models and set to High when available (GPT5, GPT5.1, o3, o3 Pro, and Gemini 3 Pro). Thinking budgets, where available, are set to max (24k for Gemini 2.5 Flash, 32k for Gemini 2.5 Pro and Opus 4.1, 64k for Sonnet 4.5 and Opus 4.5). Temperature can only be configured for Google Deepmind models. We set it to $0.7$ for Gemini 2.5 Flash and 2.5 Pro and $1.0$ for Gemini 3 Pro, as recommended in the documentation.\footnote{\href{https://ai.google.dev/gemini-api/docs/gemini-3?thinking=high}{Gemini 3 Docs}} All models are tested with web search enabled. 
\vspace{1em}

\begin{table*}[htbp]
\centering
\caption{Performance of models on the consumer activity domains in \textbf{ACE-v1-heldout}. For consistency with the leaderboard, we report the bootstrapped mean values.}
\label{tab:model-scores}
\begin{tabular}{L{4cm}|P{1.5cm}|P{1.3cm}|P{1.3cm}P{1.3cm}P{1.3cm}P{1.3cm}}
\toprule
\textbf{Model Name} & \textbf{Provider} & \textbf{Overall} & \textbf{DIY} & \textbf{Food} & \textbf{Gaming} & \textbf{Shopping} \\

\midrule
Gemini 2.5 Flash (On)      & Google    & 35.7\% & 43.7\% & 51.8\% & 28.4\% & 18.5\% \\
Gemini 2.5 Pro (On)        & Google    & 31.9\% & 40.5\% & 42.9\% & 28.5\% & 15.7\% \\
Gemini 3 Pro (High)        & Google    & 45.7\% & 44.8\% & 58.4\% & 50.9\% & 28.1\% \\
GPT 5 (High)               & OpenAI    & \textbf{56.1\%} & 55.4\% & \textbf{70.1\%} & 57.5\% & 41.7\% \\
GPT 5.1 (High)             & OpenAI    & 55.1\% & \textbf{55.8\%} & 59.1\% & 61.0\% & 44.7\% \\
o3 (On)                    & OpenAI    & 52.9\% & 52.2\% & 56.2\% & 58.5\% & 44.7\% \\
o3 Pro (On)                & OpenAI    & 55.2\% & 54.2\% & 60.2\% & \textbf{61.3\%} & \textbf{45.4\%} \\
Opus 4.1 (On)              & Anthropic & 33.8\% & 37.8\% & 46.4\% & 31.8\% & 18.8\% \\
Opus 4.5 (On)              & Anthropic & 38.3\% & 38.9\% & 45.4\% & 39.1\% & 29.5\% \\
Sonnet 4.5 (On)            & Anthropic & 35.5\% & 37.1\% & 48.3\% & 37.3\% & 19.4\% \\
\bottomrule
\end{tabular}
\end{table*}

\section{Model grading}\label{sec:lm_judges_performance}
We collected model responses eight times for each prompt. For each response, we independently score the rubric's criterion, following industry practice in using an LM judge \citep{gu2025surveyllmasajudge, zhu2025judgelmfinetunedlargelanguage}. We use Gemini 2.5 Pro with Thinking = High and Temperature set to $0.0$. Our grading methodology is hierarchical to minimize reward hacking. It involves (1) hurdles that gatekeep further rewards and (2) checking grounding for relevant criteria.
\vspace{1em}

First, for each task, we assess the prompt against the hurdle criteria. These are the most important criteria as they capture the core goal of the prompt -- such as, in Shopping, returning the requested product or, in DIY, providing a solution to the user's problem. Some criteria are phrased broadly, so without hurdles we could reward responses that are mostly irrelevant but meet a specific requirement (e.g., returning \textit{any} item under $\$50$). With a hurdle, these actions are only rewarded if the users' core goal is met. Most cases have just one hurdle criteria although some have two. On average, there are $1.32$ hurdles per case in ACE-v1-heldout.
\vspace{1em}

For DIY and Food, once the hurdle is passed, we grade as usual with a rubric -- the response is assessed for whether it meets each criterion, scoring one point for each. For example, if the criterion assesses whether the response recommends that a particular type of cleaning product is used, the response scores one point for making that recommendation. In contrast, for Gaming and Shopping, once the hurdle is passed, we assess each criterion using a three step process to check for grounding, as shown in Figure~\ref{fig:criteria_grading}. In ACE-v1-heldout, 42\% of gaming criteria and 74\% of shopping criteria require a grounding check. The intuition behind our grading methodology is that meeting criteria should be scored positively; returning nothing or failing to meet the user's core objective should score neutrally (i.e., $0$); and making up information (often called ``hallucinating'') should score negatively.
\vspace{1em}

\begin{enumerate}
\item Step one is to assess whether the content of the response meets the criterion. For example, if the criterion assesses whether the price of the returned item is below $\$100$, we check that the response states that the returned item costs less than $\$100$. If the response does not meet the criterion, it scores $0$. If it meets the criterion, we move on to step two.
\item Step two is to identify whether a grounding check is needed. This is tagged in the dataset, and applies to cases where the criterion assesses an empirical claim based on the web sources. If no grounding check is needed, the response scores \(+1\) for meeting the criterion. If a grounding check is needed, we move on to step three.
\item Step three is to check grounding. If the claim in the response being assessed is grounded in the web source, it scores \(+1\). If it is not grounded, it scores \(-1\). In Appendix~\ref{sec:appendix_technicaloverview} we describe the technical implementation of this process in more detail. 
\end{enumerate}
\vspace{1em}

Once each criterion in the rubric is graded, we compute a final score for the response by (1) linearly combining the criteria grades in the numerator and (2) counting the number of criteria in the denominator. As DIY and Food have only positive scores in the numerator, their scores have a maximum of $100\%$ and minimum of $0\%$. As Gaming and Shopping criteria can have \(-1\), \(+1\) and \(0\) values, the numerator can be negative, with a maximum of $100\%$ and theoretical minimum of $-100\%$. The theoretical minimum is only achieved if every criterion can be checked for grounding and the response hallucinates all of the required information.
\vspace{1em}

\begin{table*}[htbp]
\small
\centering
\caption{Performance of models on the criteria types in \textbf{ACE-v1-heldout}. Scores are the mean percentage of the criteria passed across the 8 runs. The values are color coded so that anything greater than 75\% is green, greater than 50\% is peach, greater than 25\% is pink and greater than 0\% is mid red. Negative values are maroon.}
\label{tab:criteria_scores}
\begin{tabular}{p{1.1cm}|p{4.9cm}| C{0.9cm} C{0.9cm} C{0.8cm} C{0.6cm} C{0.5cm} C{0.5cm} C{0.5cm} C{0.5cm} C{0.5cm} C{0.5cm}}
\toprule
\textbf{Domain} & \textbf{Criteria type} & \makecell{\textbf{Gemini}\\\textbf{2.5 Flash}\\\textbf{(On)}} & \makecell{\textbf{Gemini}\\\textbf{2.5 Pro}\\\textbf{(On)}} & \makecell{\textbf{Gemini}\\\textbf{3 Pro}\\\textbf{(High)}} & \textbf{GPT 5} & \textbf{GPT 5.1} & \textbf{o3} & \textbf{o3 Pro} & 
\makecell{\textbf{Opus}\\\textbf{4.1}} &
\makecell{\textbf{Opus}\\\textbf{4.5}} &
\makecell{\textbf{Sonnet}\\\textbf{4.5}} \\
\midrule

\multirow[t]{7}{*}{DIY} & Describes specific procedural steps & \vcolorbox{PeachPuff}{\textcolor{black}{58}} & \vcolorbox{PeachPuff}{\textcolor{black}{56}} & \vcolorbox{PeachPuff}{\textcolor{black}{56}} & \vcolorbox{PeachPuff}{\textcolor{black}{63}} & \vcolorbox{PeachPuff}{\textcolor{black}{66}} & \vcolorbox{PeachPuff}{\textcolor{black}{62}} & \vcolorbox{PeachPuff}{\textcolor{black}{63}} & \vcolorbox{LightPink}{\textcolor{black}{49}} & \vcolorbox{PeachPuff}{\textcolor{black}{50}} & \vcolorbox{LightPink}{\textcolor{black}{48}} \\
 & Other & \vcolorbox{PeachPuff}{\textcolor{black}{56}} & \vcolorbox{PeachPuff}{\textcolor{black}{56}} & \vcolorbox{PeachPuff}{\textcolor{black}{62}} & \vcolorbox{PeachPuff}{\textcolor{black}{72}} & \vcolorbox{ForestGreen}{\textcolor{black}{75}} & \vcolorbox{ForestGreen}{\textcolor{black}{75}} & \vcolorbox{ForestGreen}{\textcolor{black}{78}} & \vcolorbox{PeachPuff}{\textcolor{black}{69}} & \vcolorbox{PeachPuff}{\textcolor{black}{56}} & \vcolorbox{PeachPuff}{\textcolor{black}{74}} \\
 & Provides general DIY guidance and tips & \vcolorbox{LightPink}{\textcolor{black}{48}} & \vcolorbox{LightPink}{\textcolor{black}{41}} & \vcolorbox{LightPink}{\textcolor{black}{48}} & \vcolorbox{PeachPuff}{\textcolor{black}{61}} & \vcolorbox{PeachPuff}{\textcolor{black}{56}} & \vcolorbox{PeachPuff}{\textcolor{black}{54}} & \vcolorbox{PeachPuff}{\textcolor{black}{54}} & \vcolorbox{LightPink}{\textcolor{black}{44}} & \vcolorbox{LightPink}{\textcolor{black}{44}} & \vcolorbox{LightPink}{\textcolor{black}{41}} \\
 & Provides safety warnings & \vcolorbox{LightPink}{\textcolor{black}{44}} & \vcolorbox{LightPink}{\textcolor{black}{38}} & \vcolorbox{LightPink}{\textcolor{black}{36}} & \vcolorbox{PeachPuff}{\textcolor{black}{58}} & \vcolorbox{PeachPuff}{\textcolor{black}{54}} & \vcolorbox{PeachPuff}{\textcolor{black}{53}} & \vcolorbox{PeachPuff}{\textcolor{black}{51}} & \vcolorbox{LightPink}{\textcolor{black}{33}} & \vcolorbox{LightPink}{\textcolor{black}{31}} & \vcolorbox{LightPink}{\textcolor{black}{31}} \\
 & Provides step-by-step instructions & \vcolorbox{ForestGreen}{\textcolor{black}{88}} & \vcolorbox{ForestGreen}{\textcolor{black}{88}} & \vcolorbox{ForestGreen}{\textcolor{black}{96}} & \vcolorbox{ForestGreen}{\textcolor{black}{99}} & \vcolorbox{ForestGreen}{\textcolor{black}{100}} & \vcolorbox{ForestGreen}{\textcolor{black}{95}} & \vcolorbox{ForestGreen}{\textcolor{black}{97}} & \vcolorbox{ForestGreen}{\textcolor{black}{90}} & \vcolorbox{ForestGreen}{\textcolor{black}{93}} & \vcolorbox{ForestGreen}{\textcolor{black}{92}} \\
 & Recommends consulting a professional & \vcolorbox{LightPink}{\textcolor{black}{31}} & \vcolorbox{IndianRed}{\textcolor{white}{24}} & \vcolorbox{IndianRed}{\textcolor{white}{18}} & \vcolorbox{LightPink}{\textcolor{black}{49}} & \vcolorbox{PeachPuff}{\textcolor{black}{51}} & \vcolorbox{LightPink}{\textcolor{black}{40}} & \vcolorbox{LightPink}{\textcolor{black}{42}} & \vcolorbox{IndianRed}{\textcolor{white}{21}} & \vcolorbox{IndianRed}{\textcolor{white}{21}} & \vcolorbox{IndianRed}{\textcolor{white}{19}} \\
 & Specifies necessary materials or tools & \vcolorbox{PeachPuff}{\textcolor{black}{52}} & \vcolorbox{LightPink}{\textcolor{black}{46}} & \vcolorbox{LightPink}{\textcolor{black}{45}} & \vcolorbox{PeachPuff}{\textcolor{black}{58}} & \vcolorbox{PeachPuff}{\textcolor{black}{54}} & \vcolorbox{PeachPuff}{\textcolor{black}{52}} & \vcolorbox{PeachPuff}{\textcolor{black}{54}} & \vcolorbox{LightPink}{\textcolor{black}{42}} & \vcolorbox{LightPink}{\textcolor{black}{41}} & \vcolorbox{LightPink}{\textcolor{black}{40}} \\
\midrule

\multirow[t]{10}{*}{Food} & Meets dietary requirements & \vcolorbox{PeachPuff}{\textcolor{black}{63}} & \vcolorbox{PeachPuff}{\textcolor{black}{53}} & \vcolorbox{PeachPuff}{\textcolor{black}{69}} & \vcolorbox{PeachPuff}{\textcolor{black}{72}} & \vcolorbox{PeachPuff}{\textcolor{black}{69}} & \vcolorbox{PeachPuff}{\textcolor{black}{66}} & \vcolorbox{PeachPuff}{\textcolor{black}{69}} & \vcolorbox{PeachPuff}{\textcolor{black}{58}} & \vcolorbox{PeachPuff}{\textcolor{black}{67}} & \vcolorbox{PeachPuff}{\textcolor{black}{61}} \\
 & Meets dish feature requirements & \vcolorbox{PeachPuff}{\textcolor{black}{75}} & \vcolorbox{PeachPuff}{\textcolor{black}{70}} & \vcolorbox{ForestGreen}{\textcolor{black}{79}} & \vcolorbox{ForestGreen}{\textcolor{black}{81}} & \vcolorbox{PeachPuff}{\textcolor{black}{75}} & \vcolorbox{PeachPuff}{\textcolor{black}{73}} & \vcolorbox{ForestGreen}{\textcolor{black}{76}} & \vcolorbox{ForestGreen}{\textcolor{black}{76}} & \vcolorbox{ForestGreen}{\textcolor{black}{75}} & \vcolorbox{ForestGreen}{\textcolor{black}{76}} \\
 & Meets prep / cooking requirement & \vcolorbox{PeachPuff}{\textcolor{black}{65}} & \vcolorbox{PeachPuff}{\textcolor{black}{62}} & \vcolorbox{PeachPuff}{\textcolor{black}{72}} & \vcolorbox{PeachPuff}{\textcolor{black}{74}} & \vcolorbox{PeachPuff}{\textcolor{black}{66}} & \vcolorbox{PeachPuff}{\textcolor{black}{65}} & \vcolorbox{PeachPuff}{\textcolor{black}{71}} & \vcolorbox{PeachPuff}{\textcolor{black}{67}} & \vcolorbox{PeachPuff}{\textcolor{black}{65}} & \vcolorbox{PeachPuff}{\textcolor{black}{65}} \\
 & Meets quantity/duration requirement & \vcolorbox{ForestGreen}{\textcolor{black}{86}} & \vcolorbox{ForestGreen}{\textcolor{black}{86}} & \vcolorbox{ForestGreen}{\textcolor{black}{89}} & \vcolorbox{ForestGreen}{\textcolor{black}{93}} & \vcolorbox{ForestGreen}{\textcolor{black}{79}} & \vcolorbox{ForestGreen}{\textcolor{black}{86}} & \vcolorbox{ForestGreen}{\textcolor{black}{88}} & \vcolorbox{ForestGreen}{\textcolor{black}{87}} & \vcolorbox{ForestGreen}{\textcolor{black}{84}} & \vcolorbox{ForestGreen}{\textcolor{black}{86}} \\
 & Meets serving/portion requirement & \vcolorbox{LightPink}{\textcolor{black}{49}} & \vcolorbox{LightPink}{\textcolor{black}{33}} & \vcolorbox{LightPink}{\textcolor{black}{47}} & \vcolorbox{ForestGreen}{\textcolor{black}{83}} & \vcolorbox{PeachPuff}{\textcolor{black}{70}} & \vcolorbox{PeachPuff}{\textcolor{black}{74}} & \vcolorbox{PeachPuff}{\textcolor{black}{74}} & \vcolorbox{PeachPuff}{\textcolor{black}{52}} & \vcolorbox{LightPink}{\textcolor{black}{41}} & \vcolorbox{PeachPuff}{\textcolor{black}{56}} \\
 & Other & \vcolorbox{LightPink}{\textcolor{black}{34}} & \vcolorbox{LightPink}{\textcolor{black}{26}} & \vcolorbox{LightPink}{\textcolor{black}{28}} & \vcolorbox{LightPink}{\textcolor{black}{38}} & \vcolorbox{LightPink}{\textcolor{black}{46}} & \vcolorbox{LightPink}{\textcolor{black}{41}} & \vcolorbox{LightPink}{\textcolor{black}{40}} & \vcolorbox{IndianRed}{\textcolor{white}{20}} & \vcolorbox{LightPink}{\textcolor{black}{28}} & \vcolorbox{LightPink}{\textcolor{black}{26}} \\
 & Set list / specific recommendation & \vcolorbox{ForestGreen}{\textcolor{black}{85}} & \vcolorbox{ForestGreen}{\textcolor{black}{84}} & \vcolorbox{ForestGreen}{\textcolor{black}{84}} & \vcolorbox{ForestGreen}{\textcolor{black}{86}} & \vcolorbox{ForestGreen}{\textcolor{black}{81}} & \vcolorbox{ForestGreen}{\textcolor{black}{85}} & \vcolorbox{ForestGreen}{\textcolor{black}{91}} & \vcolorbox{ForestGreen}{\textcolor{black}{84}} & \vcolorbox{ForestGreen}{\textcolor{black}{85}} & \vcolorbox{ForestGreen}{\textcolor{black}{83}} \\
 & Provides dietary information & \vcolorbox{PeachPuff}{\textcolor{black}{51}} & \vcolorbox{LightPink}{\textcolor{black}{48}} & \vcolorbox{PeachPuff}{\textcolor{black}{54}} & \vcolorbox{PeachPuff}{\textcolor{black}{70}} & \vcolorbox{PeachPuff}{\textcolor{black}{65}} & \vcolorbox{PeachPuff}{\textcolor{black}{63}} & \vcolorbox{PeachPuff}{\textcolor{black}{65}} & \vcolorbox{PeachPuff}{\textcolor{black}{51}} & \vcolorbox{PeachPuff}{\textcolor{black}{53}} & \vcolorbox{PeachPuff}{\textcolor{black}{54}} \\
 & Provides preparation instructions & \vcolorbox{PeachPuff}{\textcolor{black}{62}} & \vcolorbox{LightPink}{\textcolor{black}{47}} & \vcolorbox{PeachPuff}{\textcolor{black}{64}} & \vcolorbox{ForestGreen}{\textcolor{black}{92}} & \vcolorbox{ForestGreen}{\textcolor{black}{84}} & \vcolorbox{PeachPuff}{\textcolor{black}{66}} & \vcolorbox{PeachPuff}{\textcolor{black}{71}} & \vcolorbox{LightPink}{\textcolor{black}{40}} & \vcolorbox{LightPink}{\textcolor{black}{36}} & \vcolorbox{LightPink}{\textcolor{black}{50}} \\
 & Provides shopping/ingredient list & \vcolorbox{PeachPuff}{\textcolor{black}{71}} & \vcolorbox{LightPink}{\textcolor{black}{50}} & \vcolorbox{ForestGreen}{\textcolor{black}{78}} & \vcolorbox{ForestGreen}{\textcolor{black}{96}} & \vcolorbox{ForestGreen}{\textcolor{black}{81}} & \vcolorbox{ForestGreen}{\textcolor{black}{77}} & \vcolorbox{ForestGreen}{\textcolor{black}{80}} & \vcolorbox{PeachPuff}{\textcolor{black}{61}} & \vcolorbox{LightPink}{\textcolor{black}{40}} & \vcolorbox{PeachPuff}{\textcolor{black}{68}} \\
\midrule

\multirow[t]{8}{*}{Gaming} & Set list / specific recommendation & \vcolorbox{LightPink}{\textcolor{black}{28}} & \vcolorbox{LightPink}{\textcolor{black}{28}} & \vcolorbox{ForestGreen}{\textcolor{black}{82}} & \vcolorbox{PeachPuff}{\textcolor{black}{64}} & \vcolorbox{PeachPuff}{\textcolor{black}{69}} & \vcolorbox{PeachPuff}{\textcolor{black}{65}} & \vcolorbox{PeachPuff}{\textcolor{black}{67}} & \vcolorbox{LightPink}{\textcolor{black}{29}} & \vcolorbox{LightPink}{\textcolor{black}{38}} & \vcolorbox{LightPink}{\textcolor{black}{39}} \\
 & Meets compatibility requirement & \vcolorbox{IndianRed}{\textcolor{white}{17}} & \vcolorbox{IndianRed}{\textcolor{white}{18}} & \vcolorbox{IndianRed}{\textcolor{white}{19}} & \vcolorbox{PeachPuff}{\textcolor{black}{51}} & \vcolorbox{PeachPuff}{\textcolor{black}{60}} & \vcolorbox{LightPink}{\textcolor{black}{25}} & \vcolorbox{LightPink}{\textcolor{black}{35}} & \vcolorbox{LightPink}{\textcolor{black}{32}} & \vcolorbox{LightPink}{\textcolor{black}{35}} & \vcolorbox{LightPink}{\textcolor{black}{25}} \\
 & Meets game/strategy requirement & \vcolorbox{LightPink}{\textcolor{black}{31}} & \vcolorbox{LightPink}{\textcolor{black}{37}} & \vcolorbox{LightPink}{\textcolor{black}{36}} & \vcolorbox{PeachPuff}{\textcolor{black}{56}} & \vcolorbox{PeachPuff}{\textcolor{black}{62}} & \vcolorbox{PeachPuff}{\textcolor{black}{55}} & \vcolorbox{PeachPuff}{\textcolor{black}{55}} & \vcolorbox{LightPink}{\textcolor{black}{49}} & \vcolorbox{PeachPuff}{\textcolor{black}{56}} & \vcolorbox{LightPink}{\textcolor{black}{46}} \\
 & Meets quantity requirement & \vcolorbox{ForestGreen}{\textcolor{black}{88}} & \vcolorbox{ForestGreen}{\textcolor{black}{89}} & \vcolorbox{ForestGreen}{\textcolor{black}{96}} & \vcolorbox{ForestGreen}{\textcolor{black}{81}} & \vcolorbox{ForestGreen}{\textcolor{black}{83}} & \vcolorbox{ForestGreen}{\textcolor{black}{88}} & \vcolorbox{ForestGreen}{\textcolor{black}{86}} & \vcolorbox{ForestGreen}{\textcolor{black}{89}} & \vcolorbox{ForestGreen}{\textcolor{black}{89}} & \vcolorbox{ForestGreen}{\textcolor{black}{77}} \\
 & Other & \vcolorbox{PeachPuff}{\textcolor{black}{55}} & \vcolorbox{LightPink}{\textcolor{black}{48}} & \vcolorbox{ForestGreen}{\textcolor{black}{80}} & \vcolorbox{PeachPuff}{\textcolor{black}{66}} & \vcolorbox{PeachPuff}{\textcolor{black}{65}} & \vcolorbox{PeachPuff}{\textcolor{black}{65}} & \vcolorbox{PeachPuff}{\textcolor{black}{68}} & \vcolorbox{LightPink}{\textcolor{black}{46}} & \vcolorbox{PeachPuff}{\textcolor{black}{52}} & \vcolorbox{PeachPuff}{\textcolor{black}{56}} \\
 & Provides game/strategy explanation & \vcolorbox{PeachPuff}{\textcolor{black}{69}} & \vcolorbox{PeachPuff}{\textcolor{black}{64}} & \vcolorbox{ForestGreen}{\textcolor{black}{79}} & \vcolorbox{PeachPuff}{\textcolor{black}{66}} & \vcolorbox{PeachPuff}{\textcolor{black}{69}} & \vcolorbox{PeachPuff}{\textcolor{black}{61}} & \vcolorbox{PeachPuff}{\textcolor{black}{61}} & \vcolorbox{PeachPuff}{\textcolor{black}{66}} & \vcolorbox{PeachPuff}{\textcolor{black}{69}} & \vcolorbox{PeachPuff}{\textcolor{black}{64}} \\
 & Provides instruction for strategy & \vcolorbox{LightPink}{\textcolor{black}{29}} & \vcolorbox{LightPink}{\textcolor{black}{29}} & \vcolorbox{PeachPuff}{\textcolor{black}{65}} & \vcolorbox{PeachPuff}{\textcolor{black}{74}} & \vcolorbox{ForestGreen}{\textcolor{black}{82}} & \vcolorbox{ForestGreen}{\textcolor{black}{79}} & \vcolorbox{ForestGreen}{\textcolor{black}{80}} & \vcolorbox{LightPink}{\textcolor{black}{47}} & \vcolorbox{PeachPuff}{\textcolor{black}{50}} & \vcolorbox{LightPink}{\textcolor{black}{41}} \\
 & Provides link(s) & \vcolorbox{Maroon}{\textcolor{white}{-5}} & \vcolorbox{Maroon}{\textcolor{white}{-2}} & \vcolorbox{Maroon}{\textcolor{white}{-0}} & \vcolorbox{PeachPuff}{\textcolor{black}{70}} & \vcolorbox{PeachPuff}{\textcolor{black}{67}} & \vcolorbox{PeachPuff}{\textcolor{black}{52}} & \vcolorbox{PeachPuff}{\textcolor{black}{61}} & \vcolorbox{IndianRed}{\textcolor{white}{24}} & \vcolorbox{LightPink}{\textcolor{black}{42}} & \vcolorbox{LightPink}{\textcolor{black}{46}} \\
\midrule

\multirow[t]{6}{*}{Shopping} & Meets pricing requirements/gives price & \vcolorbox{Maroon}{\textcolor{white}{-1}} & \vcolorbox{Maroon}{\textcolor{white}{-19}} & \vcolorbox{Maroon}{\textcolor{white}{-28}} & \vcolorbox{IndianRed}{\textcolor{white}{9}} & \vcolorbox{IndianRed}{\textcolor{white}{23}} & \vcolorbox{IndianRed}{\textcolor{white}{5}} & \vcolorbox{IndianRed}{\textcolor{white}{11}} & \vcolorbox{IndianRed}{\textcolor{white}{3}} & \vcolorbox{IndianRed}{\textcolor{white}{12}} & \vcolorbox{Maroon}{\textcolor{white}{-1}} \\
 & Meets product/vendor feature & \vcolorbox{IndianRed}{\textcolor{white}{2}} & \vcolorbox{Maroon}{\textcolor{white}{-17}} & \vcolorbox{Maroon}{\textcolor{white}{-25}} & \vcolorbox{IndianRed}{\textcolor{white}{11}} & \vcolorbox{IndianRed}{\textcolor{white}{21}} & \vcolorbox{IndianRed}{\textcolor{white}{2}} & \vcolorbox{IndianRed}{\textcolor{white}{5}} & \vcolorbox{IndianRed}{\textcolor{white}{3}} & \vcolorbox{IndianRed}{\textcolor{white}{20}} & \vcolorbox{IndianRed}{\textcolor{white}{4}} \\
 & Meets quantity requirement & \vcolorbox{ForestGreen}{\textcolor{black}{76}} & \vcolorbox{ForestGreen}{\textcolor{black}{81}} & \vcolorbox{ForestGreen}{\textcolor{black}{81}} & \vcolorbox{ForestGreen}{\textcolor{black}{79}} & \vcolorbox{ForestGreen}{\textcolor{black}{82}} & \vcolorbox{ForestGreen}{\textcolor{black}{81}} & \vcolorbox{ForestGreen}{\textcolor{black}{80}} & \vcolorbox{ForestGreen}{\textcolor{black}{75}} & \vcolorbox{PeachPuff}{\textcolor{black}{75}} & \vcolorbox{PeachPuff}{\textcolor{black}{68}} \\
 & Other & \vcolorbox{PeachPuff}{\textcolor{black}{67}} & \vcolorbox{PeachPuff}{\textcolor{black}{55}} & \vcolorbox{ForestGreen}{\textcolor{black}{78}} & \vcolorbox{ForestGreen}{\textcolor{black}{80}} & \vcolorbox{ForestGreen}{\textcolor{black}{83}} & \vcolorbox{ForestGreen}{\textcolor{black}{86}} & \vcolorbox{ForestGreen}{\textcolor{black}{82}} & \vcolorbox{PeachPuff}{\textcolor{black}{54}} & \vcolorbox{PeachPuff}{\textcolor{black}{66}} & \vcolorbox{PeachPuff}{\textcolor{black}{61}} \\
 & Set list / specific recommendation & \vcolorbox{IndianRed}{\textcolor{white}{23}} & \vcolorbox{IndianRed}{\textcolor{white}{24}} & \vcolorbox{PeachPuff}{\textcolor{black}{51}} & \vcolorbox{PeachPuff}{\textcolor{black}{66}} & \vcolorbox{PeachPuff}{\textcolor{black}{66}} & \vcolorbox{PeachPuff}{\textcolor{black}{67}} & \vcolorbox{PeachPuff}{\textcolor{black}{65}} & \vcolorbox{IndianRed}{\textcolor{white}{22}} & \vcolorbox{LightPink}{\textcolor{black}{41}} & \vcolorbox{LightPink}{\textcolor{black}{28}} \\
 & Provides link(s) & \vcolorbox{Maroon}{\textcolor{white}{-15}} & \vcolorbox{Maroon}{\textcolor{white}{-24}} & \vcolorbox{Maroon}{\textcolor{white}{-54}} & \vcolorbox{IndianRed}{\textcolor{white}{4}} & \vcolorbox{IndianRed}{\textcolor{white}{15}} & \vcolorbox{Maroon}{\textcolor{white}{-2}} & \vcolorbox{IndianRed}{\textcolor{white}{1}} & \vcolorbox{IndianRed}{\textcolor{white}{2}} & \vcolorbox{Maroon}{\textcolor{white}{-6}} & \vcolorbox{IndianRed}{\textcolor{white}{7}} \\
\bottomrule
\end{tabular}
\end{table*}

\section{Results}
We collect 8 runs from each model on each task, resulting in $32,000$ model responses (400 cases x 8 runs x 10 models). As there are more than seven criteria for each task on average, there are more than 220,000 gradings. The mean standard deviation of the scores from the 8 runs is $16.4\%$, ranging from $14.7\%$ (Opus 4.1 (Thinking = On) to $19.3\%$ (o3 (Thinking = High). This spread is due in part to enabling Thinking and non-zero temperature; and in part due to our grading methodology involving hurdles (which, if failed, make a model score $0\%$ on a task) and grounding checks (which can give negative scores to claims in responses that are not grounded in the web sources). 
\vspace{1em}

We use the mean score of the 8 runs, per model and task combination, for the leaderboard. To calculate 95\% confidence intervals, we bootstrap the data 10,000 times with a sample of 400 cases for the overall benchmark and 100 cases for the domain-specific results.\footnote{We use the bootstrapped means for the leaderboard, which vary by less than 0.1\% from the non-bootstrapped means.} See a full set of mean scores and confidence intervals in Table~\ref{tab:bootstrapped_everything} in Appendix~\ref{sec:bootstrapped_everything}. GPT 5 (Thinking = High) is the top-performing model on ACE-v1-heldout, scoring $56.1\%$, followed by o3 Pro (Thinking = On) at $55.2\%$ and GPT 5.1 (Thinking = High) at $55.1\%$. Models' mean scores for each of the four consumer domains are shown in Table~\ref{tab:model-scores}. The domains differ substantially in difficulty, leading to uneven consumer experiences in the real-world. The best performing model in Shopping scores $45.4\%$ but in DIY scores $55.8\%$, in Gaming $61.3\%$ and in Food $70.1\%$. In Food, the top-performing model has the biggest gap with the second-best model, with GPT 5 scoring 10 percentage points higher than o3 Pro (Thinking = On) at $60.2\%$.
\vspace{1em}

% \subsection{Hurdle criteria}
% Hurdle criteria are not inherently more challenging. On average, models pass the same percentage of per task hurdles as the non-hurdle criteria. However, because they are stage-gated (i.e., models score 0\% on a task if they fail the hurdle), hurdles make a substantial difference to overall scores. After accounting for hurdles, scores are on average lower by $21\%$, ranging from $-19.5\%$ for o3 Pro (Thinking = On) to $-24.3\%$ for Gemini 2.5 Pro (Thinking = On). Nonetheless, the top 5 models remain ranked in the same positions.
% \vspace{1em}

\begin{figure}[t]
\centering
\includegraphics[width=1.05\linewidth]{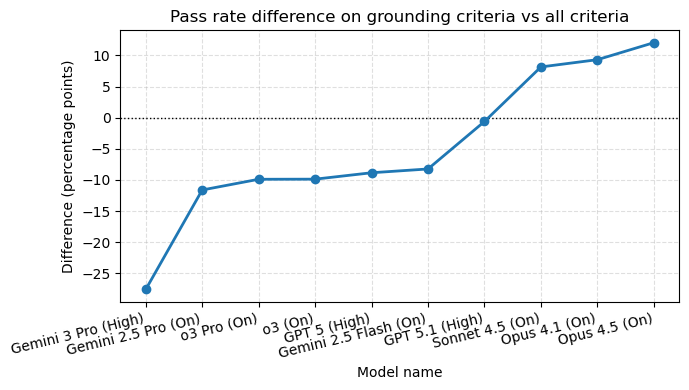}
\caption{The net difference in pass rates, comparing grounded criteria with all criteria. Negative scores indicate that models are, relatively, worse at grounding their responses than they are replying to meet the requirement of prompts.}
\label{fig:grounding_diff}
\end{figure}

\begin{table*}[b]
\small
\centering
\caption{Performance of models on \textbf{ACE-v1-heldout} compared with \textbf{ACE-v1-dev}.} %For consistency with the leaderboard, we report the bootstrapped mean values.
\label{tab:bench_os_comparison}
\begin{tabular}{L{3.8cm}|P{1.5cm}|P{1.8cm}|P{1.3cm}P{1.6cm}P{1.6cm}}
\toprule
\textbf{Model Name} & \textbf{Provider} & \textbf{Benchmark score} & \textbf{OS score} & \textbf{Score difference} & \textbf{Rank difference} \\
\midrule

Gemini 2.5 Flash (On) & Google    & 35.7\% & 40.4\% & +4.7 & 7 $\to$ 6 \\
Gemini 2.5 Pro (On)   & Google    & 31.9\% & 36.6\% & +4.7 & 10 $\to$ 10 \\
Gemini 3 Pro (High)   & Google    & 45.6\% & 47.3\% & +1.7 & 5 $\to$ 5 \\
GPT 5 (High)          & OpenAI    & \textbf{56.1\%} & 59.3\% & +3.2 & 1 $\to$ 3 \\
GPT 5.1 (High)        & OpenAI    & 55.2\% & \textbf{60.0\%} & +4.8 & 3 $\to$ 1 \\
o3 (On)               & OpenAI    & 52.9\% & 56.7\% & +3.7 & 4 $\to$ 4 \\
o3 Pro (On)           & OpenAI    & 55.2\% & 59.5\% & +4.3 & 2 $\to$ 2 \\
Opus 4.1 (On)         & Anthropic & 33.8\% & 37.6\% & +3.8 & 9 $\to$ 9 \\
Opus 4.5 (On)         & Anthropic & 38.3\% & 39.6\% & +1.3 & 6 $\to$ 8 \\
Sonnet 4.5 (On)       & Anthropic & 35.5\% & 40.0\% & +4.5 & 8 $\to$ 7 \\
\bottomrule
\end{tabular}
\end{table*}

\subsection{Grounding criteria}
Grounding criteria appear in Gaming and Shopping tasks (accounting for $42\%$ and $74\%$, respectively). We only check grounding if the response passes the main body of the criteria. Across all 8 runs, the number of grounding checks per model ranges from $2,517$ (for Opus 4.1 (Thinking = On) to $4,209$ (for o3 Pro (Thinking = On)). Models vary in how grounded they are. Gemini 3 Pro (Thinking = High) is the least grounded, passing 38.0\% of grounding tests, and GPT 5.1 (Thinking = High) is the most grounded, passing 70.8\%. These failures are more impactful than failures on non-grounding criteria as they are penalized $-1$ rather than $0$.
\vspace{1em}

We compare the pass rate of each model on all criteria versus their pass rate for grounding criteria, as shown in Figure~\ref{fig:grounding_diff}. Some models perform much worse on the grounding criteria, such as Gemini 3 Pro (Thinking = High) with a drop of $27.6$ percentage points and Gemini 2.5 Pro (Thinking = On) with a drop of $11.6$ percentage points. These models are, relatively, less grounded than they are good at creating outputs that superficially meet the prompt requirements -- and likely are hallucinating key information to appear helpful. In contrast, other models perform better on grounding criteria, such as Opus 4.5 (Thinking = On) with an increase of $12.0$ percentage points and Opus 4.1 (Thinking = On) with an increase of $9.3$ percentage points. These models are, relatively, better at grounding their responses than at just trying to meet the prompt requirements. 
\vspace{1em}

\subsection{Criteria types}
There are marked differences in how models perform on the criteria types in ACE, as shown in Table~\ref{tab:criteria_scores}. Models generally score highly on criteria that test simple aspects of responses, such as providing step-by-step instructions or meeting quantity requirements. They perform less well at more nuanced aspects of high-quality responses that require greater in-depth understanding, such as recommending to consult a professional in DIY for dangerous tasks, meeting compatibility requirements in Gaming or providing relevant dietary information in Food. Models perform poorly at providing links (both Gaming and Shopping), which are scored negatively if they are broken or hallucinated. Similarly, for Shopping, models can achieve low scores at meeting pricing requirements if the prices are hallucinated.
\vspace{1em}

\subsection{Comparison of ACE-v1-heldout and ACE-v1-dev}
We evaluated the same 10 models on the ACE leaderboard against the n=100 cases in ACE-v1-dev (available open source) to assess differences compared to ACE-v1-heldout. We use the exact same methodology (i.e., 8 runs and Gemini 2.5 Pro (Thinking = On) as a judge). Results are shown in Table~\ref{tab:bench_os_comparison}. The dataset composition is compared in Table ~\ref{tab:dataset_overview}. 
Overall, the open source dev set is similar in composition and difficulty to the benchmark. ACE-v1-dev is slightly easier, with all models performing higher than on ACE-v1-heldout. Due to the sample size, there are some differences in models' score and their rank positions. No model moves more than two rank positions and all the percentage score differences are less than 5 percentage points. Notably, GPT 5.1 (Thinking = High) replaces GPT 5 (Thinking = High) as the best performing model.
%\vspace{1em}

\section{Limitations of ACE}

\subsection{Measurement error in grounding checks}
The grounding methodology requires (1) identifying all URLs returned in the response's grounding sources and content body, (2) visiting each URL and extracting the content and (3) checking whether response claims are supported by the source. We anecdotally observed a small number of errors due to the variety of websites that models access.
\vspace{1em}

\subsection{Contamination risk}
We are open sourcing 80 cases (ACE-v1-dev) and have described the methodology behind ACE in detail in this paper. Although the heldout set used for the leaderboard remains hidden, we acknowledge the risks that greater transparency can bring. At its worst, models could climb the leaderboard without improving capabilities and creating improved experiences for consumers using AI.
\vspace{1em}

\subsection{Coverage}
We chose four domains that are high-priority for consumers and have high economic value. We are planning expansions to other domains, such as consumer finance and travel. Greater coverage will provide a more well-rounded and holistic view of the value AI creates for consumers. We also aim to update ACE with content modalities other than text-only, such as images, audio, and video.
\vspace{1em}

\subsection{Persona development}
The personas provide the model with critical context so it can assess what information to return. In real-world settings, users do not write out all of their priorities and expectations when using AI models -- yet, despite a lack of clarity in their request, they will have clear expectations for the output. These implicit expectations can be more realistically handled by feeding models multi-turn conversations where information about the users' preferences is naturally elicited.%, rather than a short persona statement. 
\vspace{1em}

\subsection{The changing Internet}
Consumer tasks often can only be executed by using web search, especially in Shopping. However, the Internet is constantly changing as new websites are created, new products launched, and new social content generated. Because the underlying reality is changing, evaluations must be refreshed and rerun to ensure they are fair. We expect ACE to be updated and rerun regularly.
\vspace{1em}

\section{Related work}
Consumer applications of AI include advances in realistic short-form video generation to creating new consumer experiences \citep{Deloitte_TheConsumerAIDossier_2025}. Increasingly, consumers use AI to research, gather and summarize information; help write and express creative expression; troubleshoot; and find shopping recommendations \citep{SommerfeldGriffin2025FiveTypesAIConsumers, Chatterji2025How, Benchek2025State}. This is translating into real economic impact. In November 2025, Adobe Analytics reported that AI-originated traffic to U.S. retail sites during Black Friday had increased 805\% compared to the previous year \citep{Reuters_BlackFriday2025_118B}. At the same time, users report serious concerns about the performance of AI Models, reporting they lack trust in accuracy, completeness, intent and data security, and are worried about hallucinations, reasoning mistakes, and privacy \citep{WEF2025ConsumersAI, McClain2025USPublicAIExperts}. A study by the World Economic Forum found that ``the most enthusiastic accelerators still demand human involvement at key moments of their buying journey'' \citep{WEF2025ConsumersAI}. It is also likely that consumer use of AI is higher than many realize -- a December 2024 report from Bain \& Co found that many consumers are not aware when they are using AI. Of 65\% of people who are self-declared ``nonusers'', 52\% were actually using generative AI-enabled tools \citep{SommerfeldGriffin2025FiveTypesAIConsumers}.
\vspace{1em}

Benchmarks measure progress in AI and, when designed carefully, help steer model training \citep{kiela-etal-2021-dynabench, schwartz2025realitychecknewevaluation, weidinger2025evaluationsciencegenerativeai}. Benchmarks are starting to measure whether models can deliver real-world value to directly benefit their users, rather than exhibiting abstract reasoning capabilities and pure ``intelligence''. For instance, the AI Productivity Index measures the ability of frontier models to perform economically valuable tasks in advanced knowledge jobs \citet{vidgen2025aiproductivityindexapex}. To-date, too little attention has been paid to benchmarking the performance of AI systems in consumer tasks. This is partly because such systems are new, and partly because consumer tasks tend to be more subjective and are often unbounded, so are intrinsically harder to benchmark fairly. A small number of evals for consumers have been released over the past year. PersonaLens assesses the personalization capabilities of models, assessing models in 20 consumer-relevant domains such as books, hotels, media and music, and shopping \citet{zhao2025personalensbenchmarkpersonalizationevaluation}. TripScore assesses whether AI models are capable of planning trips, evaluating the feasibility, reliability, and engagement of travel plans \citet{qu2025tripscorebenchmarkingrewardingrealworld}. The authors release a large-scale dataset of 4,870 queries including 219 real-world, free-form requests.

% https://arxiv.org/pdf/2506.09902
% https://arxiv.org/pdf/2510.09011
%% https://arxiv.org/pdf/2506.11763 
%% Benchmarks
%% Could add discussion of e.g., LMSYS or wildchat
% https://menlovc.com/perspective/2025-the-state-of-consumer-ai/
% https://www.bain.com/insights/understanding-the-five-types-of-ai-consumers/
% https://www.nber.org/system/files/working_papers/w34255/w34255.pdf
% https://www.weforum.org/stories/2025/01/ai-consumers-could-vs-should/
% https://www.nber.org/system/files/working_papers/w34255/w34255.pdf
% https://www.weforum.org/stories/2025/01/ai-consumers-could-vs-should/
% Similarly, a report from Pew Research in April 2025 found that 27\% of adults report using AI several times a day — yet AI experts estimate such use at 79\%. This gap could be due to expert miscalibration, but also because the public is unaware of how AI is now embedded in many widely used consumer products. 
% % https://www.pewresearch.org/internet/2025/04/03/artificial-intelligence-in-daily-life-views-and-experiences/
% found that 35\% of US adults use generative AI. They also 
%Similarly, Bain & Co found that the top reasons nonusers reported for not using generative AI as: Preferring to handle tasks themselves (34\%), Concerns about data privacy (30\%), and Not trusting the accuracy (26\%). 
% Menlo Ventures show that most users stick with one model for the majority of their interactions, favoring convenience over specialization \citet{Benchek2025State}.
% \vspace{1em}

\section{Acknowledgments}
We are very grateful to all the expert annotators who contributed to ACE. We thank everyone at Mercor who gave feedback.

\bibliography{anthology,custom}
\bibliographystyle{acl_natbib}

\appendix

\section{Technical overview of data collection}\label{sec:appendix_technicaloverview}
Model responses are collected with web search turned on. Because each provider supplies grounding information in its own format, we implement a standardization process that converts all provider-specific schemas into a single format. We extract both the URLs returned in the body of the response text and the grounding information, which includes the web links used by the model. We deduplicate and store this as a single list for each response. We then use one LM call to pull out the main claims in the model response, and a second LM call to identify the relevant links. From this we have a unified representation of claims and links.
\vspace{1em}

We use third-party services to extract relevant information from links, including a custom scraper for Reddit threads, Firecrawl for standard webpages\footnote{\href{https://www.searchapi.io/}{SearchAPI}}, and SearchAPI for YouTube video transcripts.\footnote{\href{https://www.firecrawl.dev/}{Firecrawl}} For criteria that are marked as needing a grounding check, we use the relevant source information to check whether the criteria is grounded. To prevent models from spamming recommendations and hoping that some match the criteria, ACE enforces universal standards. If multiple products are returned, all of the products must meet the requirements in each criterion -- and, where grounding is checked, all must be grounded. For instance, if a criterion checks pricing information for three products returned by a model and one of them is ungrounded (i.e., hallucinated), the response fails the grounding check for that criterion and so scores $-1$. Note that for tasks in ``Shopping'', our eval harness uses the domain-specific label for ``Shop vs. Product'' as an input when verifying purchase links. This label is provided with the ACE-v1-devset.

\section{Workflows for each domain in ACE}\label{sec:appendix_workflows}
See Table~\ref{tab:prompt_workflows}.

\begin{table*}[htbp]
\caption{Workflows for the domains in ACE, with the number and percentage of prompts assigned for \textbf{ACE-v1-heldout}. Each prompt is assigned to one and only one workflow. There are 100 cases in each domain so the counts can be interpreted as percentages.}
\label{tab:prompt_workflows}

\small 
\centering
\renewcommand{\arraystretch}{1.3}
\begin{tabular}{
    >{\centering\arraybackslash}m{2.2cm}   % Category
    >{\raggedright\arraybackslash}m{3.2cm} % Name
    >{\raggedright\arraybackslash}m{6.5cm} % Description
    >{\centering\arraybackslash}m{1.2cm}   % Number
}
\toprule
\textbf{Category} & \textbf{Workflow} & \textbf{Description} & \textbf{Number} \\

\midrule
\multirow[t]{2}{*}{DIY} & Repairs &
Tests the model’s ability to provide step-by-step instructions for home repairs.  & 65 \\
 & Crafts &
Tests the model’s ability to provide step-by-step instructions for arts and crafts projects.  & 35 \\

\midrule
\multirow[t]{3}{*}{Food}  & Meal Plan &
Tests the model’s ability to provide specific diet/meal plans based on constraints. & 37 \\

 & Potluck &
Tests the model’s ability to recommend recipes for a potluck based on a variety of circumstances and  constraints. & 38 \\

 & \makecell[l]{Cutthroat kitchen \\ Limited Resources} &
Tests the model’s ability to recommend recipes with limited available resources such as ingredients, appliances, etc. & 25 \\

\midrule
\multirow[t]{4}{*}{Gaming} & Game Design &
Tests the model’s ability to achieve a desired gameplay effect by crafting or editing game mechanics and rules.  & 13\\

 & Gaming Inspiration &
Tests the model’s ability to recommend games that are similar to a reference source such as a game review or YouTube playthrough.  & 33\\

 & Game Selection &
Tests the model’s ability to recommend games based on user preferences and constraints such as mobile vs. desktop, platforms, group play, etc. & 33 \\

 & Game Tactics &
Tests the model’s ability to provide strategic and tactical advice across various game genres. & 21 \\

\midrule
\multirow[t]{5}{*}{Shopping}  & Bargain Hunting &
Tests the model's ability to reason about product value and low-cost/bargain purchasing within significant constraints. & 15 \\

 & Compatibility &
Tests the model's ability to recommend compatible replacement parts for a given product. & 20 \\

 & Gifting &
Tests the model’s ability to recommend gifts based on the giver/recipient's preferences, relationship, etc. & 19\\

 & \makecell[l]{Profile-Based\\Recommendation}  &
Tests the model’s ability to recommend products based on a social media profile. & 19 \\

 & Vendor Recommendation &
Tests the model’s ability to recommend relevant vendors based on a desired product and availability constraints. & 27 \\

\bottomrule
\end{tabular}
\end{table*}

\section{Prompt specification text}\label{sec:appendix_prompt_specification}
The prompt specification is customized to each workflow within the domains.
See Table~\ref{tab:prompt-specification-ace}.

\begin{table*}[h!]
\caption{The prompt specifications used for the ACE leaderboard. We append these to each prompt to give the model a better chance of passing the criteria and ensure fairness between the prompt and the grading. Not all workflows have specifications. We do not have any specifications for DIY as the original prompts are specific enough.}
\label{tab:prompt-specification-ace}
\small 
\centering
\renewcommand{\arraystretch}{1.3}
\begin{tabular}{
    >{\centering\arraybackslash}m{2.2cm}   % Category
    >{\raggedright\arraybackslash}m{3.2cm} % Name
    >{\raggedright\arraybackslash}m{8.5cm} % Description
}
\toprule
\textbf{Category} & \textbf{Workflow} & \textbf{Instruction} \\

\midrule
\multirow[t]{3}{*}{Food} &
Meal Plan &
Please explicitly state whether your meal plan has each of the characteristics that I want. \\

 &
Potluck &
Please explicitly state whether each of your recommended dishes meets each of the dish feature requirements that I want. \\

 &
\makecell[l]{Cutthroat kitchen \\ Limited Resources}   &
Please explicitly state whether each of your recommended recipes has each of the characteristics that I want. \\

\midrule
\multirow[t]{3}{*}{Gaming} &
Game Design  &
Please explicitly state whether each of your recommendations meets each of the design features that I want. Include the source or purchase links, and prices if applicable. \\

 &
Game Selection &
Please explicitly state whether each of your recommendations meets each of the game features that I want. Include the source or purchase links, and prices if applicable. \\

&
Game Inspiration &
Please explicitly state whether each of your recommendations meets each of the game characteristics that I want. Include the source or purchase links, and prices if applicable. \\

 &
Game Tactics  &
Please explicitly state whether each of your recommendations meets each of the strategy features that I want. Include the source or purchase links, and prices if applicable. \\

\midrule
\multirow[t]{6}{*}{Shopping} &
Bargain Hunting  &
Please explicitly state whether each of your product recommendations meets each of the product requirements that I want. Include the source or purchase links, and prices if applicable. \\

 &
Compatibility &
Please explicitly state whether each of your product recommendations meets each of the product requirements that I want. Include the source or purchase links, and prices if applicable. \\

 % &
% Concierge &
% Please explicitly state whether each of your vendor recommendations meets each of the vendor requirements that I want. Include the source or purchase links, and prices if applicable. \\

 &
Gifting  &
Please explicitly state whether each of your product recommendations meets each of the product requirements that I want. Include the source or purchase links, and prices if applicable. \\

 &
\makecell[l]{Profile-Based\\Recommendation} &
Please explicitly state whether each of your product recommendations meets each of the product requirements that I want. Include the source or purchase links, and prices if applicable. \\

 &
Vendor Recommendation &
Please explicitly state whether each of your vendor recommendations meets each of the vendor requirements that I want. Include the source or purchase links, and prices if applicable. \\

\bottomrule
\end{tabular}
\end{table*}

\section{Bootstrapped confidence intervals for mean scores}\label{sec:bootstrapped_everything}
See Table~\ref{tab:bootstrapped_everything}.

\begin{table*}[htbp]
\small 
\centering
\renewcommand{\arraystretch}{1.25}
\caption{Bootstrapped mean scores and confidence intervals for each model and domain in ACE-v1-heldout. We draw 10,000 bootstrap samples, using 400 cases for the full benchmark and 100 cases for each domain.}
\label{tab:bootstrapped_everything}
\begin{tabular}{
    >{\centering\arraybackslash}C{3cm}   % Category
    >{\raggedright\arraybackslash}m{5cm} % Name
    >{\centering\arraybackslash}C{2cm}   % Description
    >{\centering\arraybackslash}C{2cm}   % Number
    >{\centering\arraybackslash}C{2cm}   % Percentage
}
\toprule
\textbf{Domain} & \textbf{Model} &\textbf{Mean (\%)} & \textbf{CI Lower (\%)} & \textbf{CI Upper (\%)} \\
\midrule
\multirow[t]{10}{*}{Overall} & Gemini 2.5 Flash (On) & 35.7 & 32.8 & 38.6 \\
 & Gemini 2.5 Pro (On) & 31.9 & 29.3 & 34.7 \\
 & Gemini 3 Pro (High) & 45.7 & 42.7 & 48.6 \\
 & GPT 5 (High) & 56.1 & 52.8 & 59.4 \\
 & GPT 5.1 (High) & 55.1 & 51.9 & 58.3 \\
 & o3 (On) & 52.9 & 49.8 & 56.0 \\
 & o3 Pro (On) & 55.2 & 52.1 & 58.5 \\
 & Opus 4.1 (On) & 33.8 & 31.0 & 36.6 \\
 & Opus 4.5 (On) & 38.3 & 35.3 & 41.3 \\
 & Sonnet 4.5 (On) & 35.5 & 32.5 & 38.4 \\
\midrule
\multirow[t]{10}{*}{DIY} & Gemini 2.5 Flash (On) & 43.7 & 38.2 & 49.0 \\
 & Gemini 2.5 Pro (On) & 40.5 & 35.3 & 45.7 \\
 & Gemini 3 Pro (High) & 44.8 & 39.8 & 49.8 \\
 & GPT 5 (High) & 55.4 & 50.0 & 60.7 \\
 & GPT 5.1 (High) & 55.8 & 50.6 & 60.8 \\
 & o3 (On) & 52.2 & 47.1 & 57.1 \\
 & o3 Pro (On) & 54.2 & 49.2 & 59.0 \\
 & Opus 4.1 (On) & 37.8 & 33.0 & 42.7 \\
 & Opus 4.5 (On) & 38.9 & 33.7 & 43.9 \\
 & Sonnet 4.5 (On) & 37.1 & 32.1 & 42.2 \\
\midrule
\multirow[t]{10}{*}{Food}  & Gemini 2.5 Flash (On) & 51.8 & 46.7 & 56.7 \\
 & Gemini 2.5 Pro (On) & 42.9 & 38.1 & 47.5 \\
 & Gemini 3 Pro (High) & 58.4 & 53.4 & 63.0 \\
 & GPT 5 (High) & 70.1 & 64.5 & 75.3 \\
 & GPT 5.1 (High) & 59.1 & 52.5 & 65.6 \\
 & o3 (On) & 56.2 & 50.5 & 62.1 \\
 & o3 Pro (On) & 60.2 & 53.9 & 66.0 \\
 & Opus 4.1 (On) & 46.4 & 41.3 & 51.6 \\
 & Opus 4.5 (On) & 45.4 & 40.3 & 50.5 \\
 & Sonnet 4.5 (On) & 48.3 & 42.8 & 53.6 \\
\midrule
\multirow[t]{10}{*}{Gaming}  & Gemini 2.5 Flash (On) & 28.4 & 22.5 & 34.3 \\
 & Gemini 2.5 Pro (On) & 28.5 & 22.6 & 34.7 \\
 & Gemini 3 Pro (High) & 50.9 & 44.8 & 57.2 \\
 & GPT 5 (High) & 57.5 & 50.6 & 64.0 \\
 & GPT 5.1 (High) & 61.0 & 54.6 & 67.3 \\
 & o3 (On) & 58.5 & 52.0 & 64.6 \\
 & o3 Pro (On) & 61.3 & 54.8 & 67.2 \\
 & Opus 4.1 (On) & 31.8 & 26.2 & 37.8 \\
 & Opus 4.5 (On) & 39.1 & 32.9 & 45.7 \\
 & Sonnet 4.5 (On) & 37.3 & 31.1 & 43.7 \\
\midrule
\multirow[t]{10}{*}{Shopping} & Gemini 2.5 Flash (On) & 18.5 & 13.9 & 23.2 \\
 & Gemini 2.5 Pro (On) & 15.7 & 11.7 & 20.2 \\
 & Gemini 3 Pro (High) & 28.1 & 22.0 & 34.6 \\
 & GPT 5 (High) & 41.7 & 34.0 & 48.9 \\
 & GPT 5.1 (High) & 44.7 & 37.4 & 51.7 \\
 & o3 (On) & 44.7 & 37.6 & 51.7 \\
 & o3 Pro (On) & 45.4 & 38.2 & 53.0 \\
 & Opus 4.1 (On) & 18.8 & 14.3 & 23.8 \\
 & Opus 4.5 (On) & 29.5 & 23.1 & 36.1 \\
 & Sonnet 4.5 (On) & 19.4 & 14.3 & 24.6 \\
\bottomrule
\end{tabular}
\end{table*}

\end{document}